\documentclass[conference]{IEEEtran}
\usepackage{amsmath,amssymb,amsfonts,mathrsfs}
\usepackage{textcomp}
\def\BibTeX{{\rm B\kern-.05em{\sc i\kern-.025em b}\kern-.08em
    T\kern-.1667em\lower.7ex\hbox{E}\kern-.125emX}}
\usepackage{soul}
\usepackage[utf8]{inputenc}
\usepackage{comment}
\usepackage{balance}
\usepackage{multirow}
\usepackage{pifont}
\usepackage{xifthen}
\usepackage[underline=false]{pgf-umlsd}
\usepackage{tikz}
\usepackage{algorithm}
\usepackage[noend]{algpseudocode}
\usepackage{algpseudocode}
\newcommand{\formatcomment}[1]{\scriptsize\textcolor{blue!25!black}{\texttt{#1}}}

\algrenewcommand{\algorithmiccomment}[1]{\hfill\parbox{5.2cm}{\formatcomment{//\,#1}}}
\algrenewcommand\algorithmicprocedure{\textbf{algorithm}}
\algnewcommand\algorithmicforeach{\textbf{for each}}
\algdef{S}[FOR]{ForEach}[1]{\algorithmicforeach\ #1\ \algorithmicdo}
\usepackage{enumitem}
\usepackage[n ,advantage, operators, sets , adversary, landau, probability, notions, logic, ff , mm, primitives, events, complexity, asymptotics, keys]{cryptocode}
\usepackage{booktabs} 
\usepackage{comment}
\usepackage[normalem]{ulem} 
 \newcommand{\superscript}[1]{\ensuremath{{}^{\textrm{\scriptsize #1}}}}
 
 \newcommand{\mntext}[1]{\colorbox{SkyBlue}{\begin{color}{black}#1\end{color}}}
 \newcommand{\mn}[2][]{{\tiny\superscript{\mntext{\arabic{mn}}}}\marginpar{\scriptsize{
 			\ifthenelse{\isempty{#1}}
 			{\mntext{\parbox{0.95\marginparwidth}{\superscript{\arabic{mn}}~\raggedright{#2}}}}
 			{\mntext{\parbox{0.95\marginparwidth}{\superscript{\arabic{mn}}#1 says~:~\raggedright{#2}}}}
 		}}\stepcounter{mn}}

\usepackage{mathtools}
\usepackage{graphicx}  
\usepackage{xcolor}
\usepackage{tikz}
\usepackage{pgfplots}
\usetikzlibrary{decorations.pathreplacing,calc}

\usepackage{natbib}
\usepackage{array}
\usepackage{siunitx}
\usepackage{tkz-euclide,subfig}

\definecolor{Purple}{rgb}{128, 0, 128}

\makeatletter
\newcommand{\newlineauthors}{%
  \end{@IEEEauthorhalign}\hfill\mbox{}\par
  \mbox{}\hfill\begin{@IEEEauthorhalign}
}
\makeatother
\usepackage{url}

\begin{document}

\title{
MIDST Challenge at SaTML 2025: \\Membership Inference over Diffusion-models-based Synthetic Tabular data}

\author{
\IEEEauthorblockN{Masoumeh Shafieinejad\textsuperscript{1}, Xi He\textsuperscript{1,2}, Mahshid Alinoori\textsuperscript{1}, John Jewell\textsuperscript{1}, \\Sana Ayromlou\textsuperscript{1}, Wei Pang\textsuperscript{2}, Veronica Chatrath\textsuperscript{1}, Gauri Sharma\textsuperscript{3}, Deval Pandya\textsuperscript{1}}
\IEEEauthorblockA{\textit{\textsuperscript{1}Vector Institute, \textsuperscript{2}University of Waterloo, \textsuperscript{5}McGill University}\\
\{masoumeh, xi.he, mahshid.alinoori, deval.pandya\}@vectorinstitute.ai}  johnjewelldev@gmail.com, sayromlou@gmail.com, w3pang@uwaterloo.ca, \\veronica.chatrath@robotics.utias.utoronto.ca, gauri.sharma@mail.mcgill.ca
}

\maketitle

\begin{abstract}
Synthetic data is often perceived as a silver-bullet solution to data anonymization and privacy-preserving data publishing. Drawn from generative models like diffusion models, synthetic data is expected to preserve the statistical properties of the original dataset while remaining resilient to privacy attacks. 
Recent developments of diffusion models have been effective on a wide range of data types, but their privacy resilience, particularly for tabular formats, remains largely unexplored. MIDST challenge sought a quantitative evaluation of the privacy gain of synthetic tabular data generated by diffusion models, with a specific focus on its resistance to membership inference attacks (MIAs). Given the heterogeneity and complexity of tabular data, multiple target models were explored for MIAs, including diffusion models for single tables of mixed data types and multi-relational tables with interconnected constraints. MIDST inspired the development of novel black-box and white-box MIAs tailored to these target diffusion models as a key outcome, enabling a comprehensive evaluation of their privacy efficacy. The MIDST GitHub repository is available at:  \url{https://github.com/VectorInstitute/MIDST}

\end{abstract}

\section{Introduction}
Privacy regulations across the globe ---  European General Data
Protection Regulation (GDPR), the Canadian Personal Information Protection and Electronic Documents Act (PIPEDA), California’s Consumer Privacy Act (CCPA), etc --- call for data anonymization as a main privacy principle. Industries are exploring privacy-preserving technologies to comply. JP Morgan has published an extensive report \cite{potluru2024synthetic} stating ``for a highly regulated finance industry, synthetic data is a potential approach for dealing with issues related to privacy, fairness, and explainability''. Similarly, the US Department of Health and Human Services regards synthetic data as a powerful tool that can address main challenges in making health care data accessible~\cite{Gonzales2023Synthetic}. Vector Institute, a national AI institute in Canada, organized a boot camp for industry sponsors to 
promote diffusion models for synthesizing tabular data \cite{vector2025DMbootcamp}. However, without effective measures against privacy attacks, true anonymity may not be achieved. Vector Institute, organized a membership inference attack challenge on synthetic tabular data generated by diffusion models, named MIDST challenge, as part of the IEEE SaTML 2025 conference. This competition is a contribution to the broader, real-world challenge of translating industry and regulatory risk assessments into technical terms.

\subsection{Novelty} 
The usage of membership inference attacks for anonymity measurement is not new. However, their application on complex tabular data generated by diffusion models had not yet been addressed by either the research community or the industry prior to MIDST. MIDST also extended this evaluation to multi-table data synthesis, a popular application in industry, which had not been addressed for evaluation in research before, even outside the diffusion model context. A relevant membership inference competition was held by Microsoft in SaTML 2023, MICO project \cite{microsoft2023mico}, evaluating the effectiveness of differentially private model training as a mitigation against white-box membership inference attacks. However, MIDST focuses on complex tabular data and evaluates the privacy efficacy or limitations of synthetic data generators. 

\section{Background}
\subsection{Tabular Data Synthesis}
Synthetic data has attracted significant interest for its ability to tackle key challenges in accessing high-quality training datasets. These challenges include: i) privacy \cite{assefa2020generating,hernandez2022synthetic}, ii) bias and fairness \cite{Breugel2023Beyond}, and iii) data scarcity \cite{fonseca2023tabular, zheng2022diffusion}. The interest in synthetic data has extended to various commercial settings, notably in healthcare \cite{Gonzales2023Synthetic} and finance \cite{potluru2024synthetic} sectors. The synthesis of tabular data, among all data modalities, is a critical task with approximately 79\% of data scientists working with it on a daily basis \cite{Breugel2023Can}.

\subsection{Diffusion Models}
Diffusion models have emerged as powerful tools for data synthesis, demonstrating remarkable success in various domains~\cite{rombach2022high}. These models are particularly noted for their strong capabilities in controlled (conditioned) generation. They have been used for generating synthetic data in both unconditional settings, for single tables, \cite{kotelnikov2023tabddpm, zhang2023mixed, Lee2023CoDi, kim2022stasy} and  \cite{pang2024clavaddpmm, Liu2024Controllable} conditioned setting for multiple interconnected tables. 

\subsection{Membership Inference Attacks (MIAs)}
In tabular synthesis literature, the generative model is commonly evaluated by a basic distance to the closest record (DCR) metric for its privacy protection. However, the generated synthetic data indicates its privacy deficiencies when assessed by stronger privacy measurements such as membership inference attacks \cite{Stadler2022Synthetic, giomi2022unified}. The evaluation of diffusion models by membership inference attacks was largely limited to computer vision \cite{Breugel2023Membership, Duan2023AreDiff, Carlini2023Extracting}, prior to the MIDST competition. With the success of diffusion models in generating tabular synthetic data, MIDST highlighted the need to evaluate the models with proper privacy metrics.

\section{MIDST Challenge Design}
The generative models are developed on the training data set to generate synthetic data. They are expected to learn the statistics without memorizing the individual data. To evaluate this promise, membership inference attacks assess whether the model distinguishes between the \textit{training data set} and a \textit{holdout data set}, both are derived from the same, larger data set. For each of the four tasks, a set of models were trained on different splits of a public dataset. For each of these models, $m$ challenge points were provided; exactly half of which are members (i.e., used to train the model) and half are non-members (i.e., from the holdout set; they come from the same public dataset as the training set, but were not used to train the model). The goal of the participants is to determine which challenge points are members and which are non-members.

\subsection{Challenge Tracks}
MIDST challenge was composed of four different tracks, each associated with a separate category. The categories are defined based on the access to the generative models and the type of the tabular data as follows:
\begin{enumerate}
    \item Access to the models: black-box, Data: single table
    \item Access to the models: white-box, Data: single table
    \item Access to the models: black-box, Data: multi-table
    \item Access to the models: white-box, Data: multi-table
\end{enumerate}
In white-box attacks, the participants had access to the models and their generated synthetic output. Training sets for these models are selected from a public dataset. In black-box attacks, the access to the same information as the white-box attack was granted, except for the models. 

\subsection{Models and Datasets}
Vector Institute held a boot camp on diffusion-model-based tabular synthesis used for a variety of applications for industry stakeholders on September 3-5, 2024 \cite{vector2025DMbootcamp}. The plug-and-play diffusion-model-based reference implementations include: TabDDPM (single table)~\cite{kotelnikov2023tabddpm}, TabSyn (single table)~\cite{zhang2023mixed}, and ClavaDDPM (multi-table)~\cite{pang2024clavaddpmm}. The repository of the models and the utilized datasets are publicly available through the link: \url{https://github.com/VectorInstitute/MIDSTModels}. The same models were included in the MIDST challenge for privacy evaluation. To facilitate the participation of all interested researchers regardless of their computing capabilities, MIDST also provided a set of 30 shadow models for each of these three models. The shadow models were the same for black-box and white-box tasks. The participants were free to choose these shadow models and/or generate their own if needed in developing their MIAs. These models were each trained on a 20,000-record subsample of the \textit{Transaction}\footnote{For multi-table tracks, the corresponding records from the other tables in the Berka dataset were also included.} table from the publicly available Berka dataset \cite{berka2000guide}. The Berka dataset is a collection of 8 tables representing financial information from a Czech bank. The dataset deals with over 5,300 bank clients with approximately 1,000,000 transactions. Additionally, the bank represented in the dataset has extended close to 700 loans and issued nearly 900 credit cards.

\begin{table*}
\centering
\begin{tabular}{l l c l c}
\toprule
Track & Winner & Success & Runner-up & Success \\
\midrule
White-box Single Table & Tartan Federer~\cite{wu2025winning} & 46\% & Yan Pang~\cite{pang2025solution} & 39\% \\
White-box Multi-table & Tartan Federer & 35\% & ** & ** \\
Black-box Single Table & Tartan Federer & 25\% & CITADEL \& UQAM~\cite{Ensemble} & 22\% \\
Black-box Multi-table & Tartan Federer & 23\% & Cyber@BGU~\cite{german2025miaept} & 20\% \\
\bottomrule
\end{tabular}
\caption{Competition results across tracks. **We received several submissions for the white-box multi-table task; however, their performance did not significantly exceed that of random guessing.}
\label{tab:competition_results}
\end{table*}

\subsection{Submission, Evaluation and Scoring}
MIDST included three phases: train, dev, and final. For 30 shadow models in train (for each model in each track), the full training dataset, and consequently the ground truth membership data for challenge points were revealed. The participants could use these [shadow] models to develop their attacks. For 20 models in the dev and 20 models in final sets, no ground truth was revealed and participants needed to submit their membership predictions for challenge points. During the competition, a live scoreboard showed the results on the dev challenges. The final ranking was decided on the final set; scoring for this dataset was withheld until the end of the competition. 

Submissions were ranked based on their performance in membership inference against the associated models. For each challenge point, the submission provided a value, indicating the confidence level with which the challenge point is a member. Each value being a floating point number in the range [0.0, 1.0], with 1.0 indicating certainty that the challenge point is a member, and 0.0 indicating certainty that it is a non-member. Submissions were evaluated according to their True Positive Rate at 10\% False Positive Rate (i.e. TPR \@ 0.1 FPR). In this context, positive challenge points were members and negative challenge points were non-members. For each submission, the scoring program concatenated the confidence values for all models (dev and final treated separately) and compared these to the reference ground truth. The scoring program determined the minimum confidence threshold for membership such that at most 10\% of the non-member challenge points are incorrectly classified as members. The score captured the True Positive Rate achieved by this threshold (i.e., the proportion of correctly classified member challenge points). The live scoreboard showed additional scores (i.e., TPR at other FPRs, membership inference advantage, accuracy, AUC-ROC score), but these are only informational.


\subsection{The competition timeline}
The models and dataset were published by December $1^{st}$, 2024. The submission to the live scoreboard was open from December $1^{st}$, 2024 until February $27^{th}$, 2025. The submissions for the final phase were collected on February $28^{th}$. 
The winners of all tasks were announced to the competition participates on March $10^{th}$, 2025, as well as officially during SaTML conference, April 9-11, in Copenhagen. 


\section{Results}
We received over 700 submissions from 71 participants across four tracks. The Tartan Federer team placed first in all tracks, and we announced runner-ups in three of the four tracks. For the remaining track, white-box multi-table, we did not announce a runner-up, as the submitted approaches did not significantly outperform random guessing.

\subsection{Winning solutions}
The MIDST results are provided in Table~\ref{tab:competition_results}. 
In the white-box track, our top-performing teams used different approaches in their attack development. Tartan Federer \cite{wu2025winning} used SecMI \cite{Duan2023AreDiff} as a starting point for their attack design. While SecMI has shown success in image-based diffusion models, its original design proved less effective for tabular data – highlighting the effect of data domain on attack development. Tartan Federer  identified noise initialization as a key factor influencing attack efficacy and proposed a machine-learning-driven approach that leverages loss features across different noises and time steps. Inspired by the success of their GSA approach in computer vision \cite{pang2025whiteboxMIA}, Yan Pang leveraged the differences in gradients between member and non-member samples for their attack development \cite{pang2025solution}.
In the black-box track, the best performing submissions employed a diverse set of techniques too. Cyber\@BGU team \cite{german2025miaept} leveraged shadow models, auxiliary machine learning models, and an attack classifier to craft their attack. Tartan Federer also used shadow model parameters for their attack development. CITADEL \& UQAM \cite{Ensemble} performed their MIA through an ensemble technique. In addition to shadow-model-based predictions of RMIA \cite{Meeus2025RMIA} and DOMIAS \cite{Breugel2023MIAagainst} their meta-classifier takes continuous features of the data as well as several measurements of Gower distance between the data points and the synthetic dataset as inputs.

\subsection{Directions for future research}
\subsubsection{TabSYN vs TabDDPM}
MIDST provided two models with different structures for single table tracks: TabSYN and TabDDPM. The competition considers the highest score achieved in attacking either of the models for ranking. Most of the attacks submitted targeted TabDDPM, a few that attacked both achieved higher scores for TabDDPM. It remains an open question whether the preference comes from the fact that latent space diffusion models like TabSYN are less explored, or that the structure makes these models more resilient against membership inference attacks.Evidence of the former argument is the SecMI attack, where latent space diffusion models are considered in attack extension rather than in the default design of the attack itself.

\subsubsection{Single table vs multi-table}
MIDST uses Transaction table from the Berka dataset for the single table tracks. For multi-table tracks, the other tables from the Berka dataset are added as well. However, the MIDST challenge points for all tracks were restricted to the Transaction table. An intuitive consequence from this setup would be that the attacks designed for single table models are applicable to multi-table ones, with similar success rate if they opt to not use the additional information from the other tables, and higher success rates if they opt to do so. However, the submitted results – particularly on white-box track, do not follow this intuition.

\subsubsection{Comparison with the other AI-Gen for tabular synthesis} 
Diffusion models perform exceptionally well for tabular synthesis. MIDST results show that this synthesis is not free of privacy leakage. However, without further investigation and comparison with other GenAI approaches, it remains unclear whether this privacy leakage is specific to diffusion models.

\section{Conclusion}
MIDST challenge represents a milestone in assessing the privacy limitations of diffusion-based synthetic tabular data. The competition yielded novel membership inference attacks that expose the vulnerabilities of diffusion models, underscoring that synthetic generation is not necessarily a default privacy guarantee. These results highlight the urgent need for better assessments and audits of the privacy risks  in the life cycle of synthetic tabular data.

\section{Acknowledgment}
We are deeply grateful to the University of Waterloo Cybersecurity \& Privacy Institute (CPI) and the Data Systems Group (DSG) for their generous sponsorship of the MIDST competition. Also, we'd like to thank MICO organizers, for their open source project, and very helpful comments.

\bibliographystyle{abbrv}

\end{document}